%% file: gwcl.tex
\documentclass[10pt,twocolumn,letterpaper]{article}
\usepackage[utf8]{inputenc} 
\usepackage[T1]{fontenc}   
\usepackage{iccv}              


\definecolor{iccvblue}{rgb}{0.21,0.49,0.74}
\usepackage[pagebackref,breaklinks,colorlinks,allcolors=iccvblue]{hyperref}

\usepackage{caption}
\usepackage{subcaption}
\usepackage{graphicx} 
\usepackage[american]{babel}
\usepackage{microtype}
\usepackage{multirow}
\usepackage{booktabs}
\usepackage{algorithm}

\usepackage{algorithm}
\usepackage{algorithmicx}
\usepackage{algpseudocode}

\usepackage{bm}

\usepackage{wrapfig}
\usepackage{pifont}
\usepackage{multirow}
\usepackage[export]{adjustbox}
\usepackage{color}
\usepackage{colortbl}
\usepackage{lipsum}
\usepackage{pifont}       
\usepackage{bbding}    
\usepackage[dvipsnames]{xcolor}
\usepackage{array}

\usepackage[capitalize]{cleveref}
\usepackage{float}
\crefname{section}{Sec.}{Secs.}
\Crefname{section}{Section}{Sections}
\Crefname{table}{Table}{Tables}
\crefname{table}{Tab.}{Tabs.}
\definecolor{red}{RGB}{255,0,0}
\definecolor{blue}{RGB}{0,0,255}
\definecolor{green}{RGB}{0,255,0}
\definecolor{mygray}{gray}{.9}
\definecolor{mygray2}{gray}{.5}
\definecolor{mywarning}{RGB}{233,144,61}
\definecolor{mygreen}{RGB}{93,174,86}
\definecolor{codefunc}{RGB}{73,122,234}

\definecolor{mygreen}{RGB}{0,154,85}
\definecolor{myy}{RGB}{126,95,0}
\definecolor{myred}{RGB}{212,121,116}
\definecolor{myblue}{RGB}{184, 134, 73}
\definecolor{mynewgreen}{RGB}{113,188,169}
\definecolor{mypurple}{RGB}{123,104,238}
\colorlet{R1}{myblue}
\colorlet{R2}{mypurple}
\colorlet{R3}{myred}
\colorlet{R6}{mypurple}
\definecolor{mycite}{RGB}{73,123,184}
\colorlet{cite}{mycite}

\newcommand{\x}{{\bm{x}}}

\newcommand{\z}{{\bm{z}}}
\newcommand{\h}{{\bm{h}}}

\def\paperID{5263} 
\def\confName{ICCV}
\def\confYear{2025}

\title{Graph-Weighted Contrastive Learning\\
	 for Semi-Supervised Hyperspectral Image Classification}

\author{Yuqing Zhang$^1$, Qi Han$^1$, Ligeng Wang$^1$, Kai Cheng$^1$, Bo Wang$^{2,3}$, and Kun Zhan$^{1,2,\star}$\\
	1. School of Information Science and Engineering, Lanzhou University\\
	2. Key Laboratory of AI and Information Processing, Hechi University\\
	3. School of Artificial Intelligence and Smart Manufacturing, Hechi University\\
	{\small \url{https://github.com/kunzhan/semiHSI}}}

\begin{document}
\maketitle
\begin{abstract}
Most existing graph-based semi-supervised hyperspectral image classification methods rely on superpixel partitioning techniques. However, they suffer from misclassification of certain pixels due to inaccuracies in superpixel boundaries, \ie, the initial inaccuracies in superpixel partitioning limit overall classification performance. In this paper, we propose a novel graph-weighted contrastive learning approach that avoids the use of superpixel partitioning and directly employs neural networks to learn hyperspectral image representation. Furthermore, while many approaches require all graph nodes to be available during training, our approach supports mini-batch training by processing only a subset of nodes at a time, reducing computational complexity and improving generalization to unseen nodes. Experimental results on three widely-used datasets demonstrate the effectiveness of the proposed approach compared to baselines relying on superpixel partitioning.
\end{abstract}
\section{Introduction}Graph transductive learning~\cite{4305352,NIPS2003_87682805}, based on graph spectral theory~\cite{chung1997spectral}, offers strong transductive capability. Due to the graph transductive capability in semi-supervised learning, many graph-based methods~\cite{4305352,SGL,SLGConv,ConGCN,CAD_GCN,MSSGU,DMSGer,GIG,Nonlocal} have been developed for semi-supervised hyperspectral image (HSI) classification.

\begin{figure}[!t]\centering
	\includegraphics[width=0.48\textwidth]{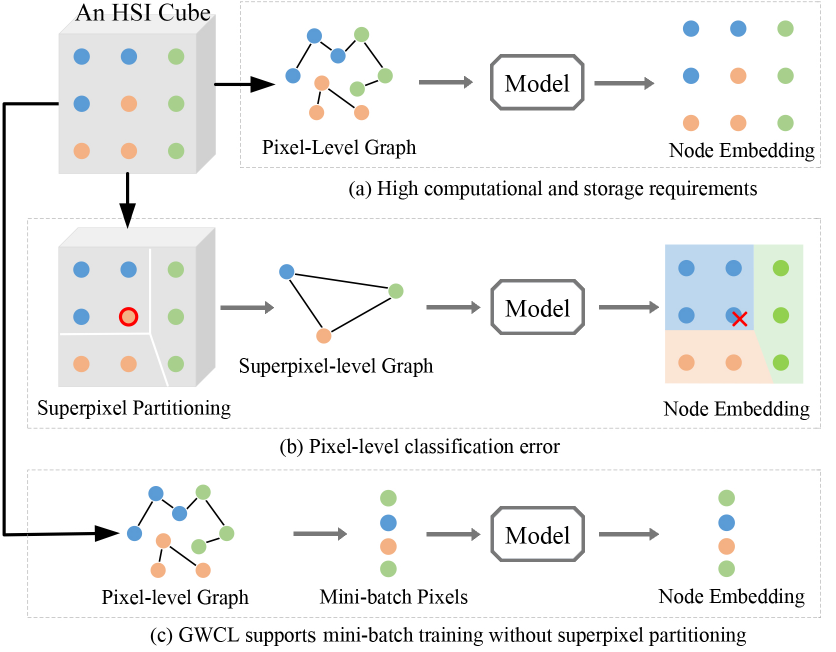}
	\caption{Different approaches to applying graphs in semi-supervised HSI classification include: (a) Pixel-level graph has high computational and storage requirement; (b) Superpixel-level graph often suffers from pixel misclassification due to inaccuracies in superpixel partitioning; (c) Our proposed GWCL method eliminates the need for superpixel partitioning and supports mini-batch training, offering a more efficient and accurate solution.}\label{motivation}
\end{figure}

Existing graph-based semi-supervised HSI classification algorithms\cite{wu2024hyperspectral} are subject to three fundamental limitations: (1) Constructing graphs with image pixels as nodes is challenging due to the computational and storage requirements of large-scale graph matrices; (2) consequently, graphs are often built using superpixels, which limits performance as misclassified pixels can be grouped within the same superpixel; and (3) implementing graph-based HSI classification algorithms in mini-batch training is difficult, as they typically require all nodes to be processed together. Intuitively, the most straightforward approach~\cite{Nonlocal,miniGCN} is to construct the graph using HSI pixels as nodes, as shown in Fig.~\ref{motivation}(a). However, HSI contains a large number of pixels, which presents significant challenges for training neural networks. Hong~\etal~\cite{miniGCN} addressed this issue by proposing miniGCN, which samples subgraphs from the complete graph to enable mini-batch training. However, their graph construction approach relies solely on pixelwise spectral information to calculate edge weights, overlooking the pixel-spatial information that is crucial for accurate HSI classification.
In recent years, there has been a growing trend toward constructing smaller graphs at the superpixel level for HSI classification~\cite{GIG,MView,ConGCN,MSSGU,DMSGer,SGL,SLGConv,CAD_GCN}, as shown in Fig.~\ref{motivation}(b). This involves first dividing the HSI into multiple superpixels, with superpixel node features being updated during training. The most commonly used superpixel segmentation algorithms include Simple Linear Iterative Clustering (SLIC)~\cite{SLIC} and Entropy Rate Superpixel (ERS)~\cite{ERS}.  Yu~\etal~\cite{MView} used SLIC~\cite{SLIC} to roughly divide HSI into many superpixels to reduce the computational burden. Jiang~\etal~\cite{MSSG} constructed a multi-scale spectral-spatial graph based on ERS~\cite{ERS}. In addition, Wan~\etal~\cite{CAD_GCN} used a superpixel-induced graph, which is transformed from the original 2D image grids.
In HSI classification, finer image partitioning leads to smaller superpixel sizes, which are more conducive to accurate pixel-level classification.
Consequently, HSI is often segmented into thousands of superpixels. However, since adjacent pixels from different classes are grouped into the same superpixel, pixel-wise missegmentation is inevitable, which constrains classification performance.

Considering the limitations of existing graph-based semi-supervised HSI classification methods, we propose a novel graph-weighted contrastive learning (GWCL) approach that eliminates the reliance on superpixel partitioning and directly utilizes neural networks to learn HSI representations, as illustrated in Fig.~\ref{motivation}(c). At the core of our approach is a newly defined graph-weighted contrastive loss, designed to enhance pixel-wise feature learning. This loss is derived from maximizing the lower bound of mutual information, which captures nonlinear statistical dependencies between variables, serving as a measure of true relationships. To boost classification performance, our approach encourages the encoder to learn features that highlight similarities within the same class while distinguishing between different classes. By maximizing mutual information, we enhance the model's ability to learn pair-wise similarity between positive pairs~\cite{CPC,MINE,DSSN2023,infomatch}, enabling the use of rich supervisory signals from large volumes of unlabeled pixels.
In our approach, we assign pair-wise weights based on the graph structure, using these weights to determine the similarity between positive pairs, as pixels connected by an edge are considered positive pairs. These weights quantify the similarity between pixel representations, effectively capturing the probability that they belong to the same class. By employing this soft, pair-wise weight construction method, the model becomes less sensitive to low-quality positive pairs, relying more on high-quality pairs, which significantly enhances its overall performance.

Additionally, to fully leverage both pixel-spectral and pixel-spatial information, we construct a global graph with pixels as nodes. Recognizing that traditional methods often overlook pixel-spatial information in graph construction, our approach incorporates both pixel-spectral and pixel-spatial similarities when calculating edge strengths. Given the computational challenges associated with pixel-level graph matrices, our approach is designed to support mini-batch training, enabling the use of the graph matrix during training while optimizing computational resource usage. GWCL processes pixels directly throughout the entire process, without relying on superpixel partitioning. This approach not only reduces computational costs but also effectively applies the graph matrix during training. Extensive experiments demonstrate that GWCL outperforms a set of related methods, showcasing its superior performance.

The contributions of our approach are summarized as follows.
\begin{itemize}
	\item We introduce a novel graph-weighted contrastive learning method for semi-supervised HSI classification, featuring a newly defined graph-weighted contrastive loss derived from maximizing the lower bound of mutual information.
	\item Our approach supports mini-batch training and inference by processing only a subset of nodes, whereas most existing methods require all nodes from the graph to be present during training. This not only reduces computational complexity but also facilitates the use of pair-wise graph weights during the training process.
	\item Experiments on three widely used datasets demonstrate that, under the same selection criteria for labeled pixels, GWCL outperforms methods based on superpixel-level graph HSI classification baselines.
\end{itemize}

The rest of this paper is organized as follows. In Section \ref{related}, we review related work on graph-based methods for HSI classification. Our proposed approach, Graph-Weighted Contrastive Learning, is introduced in Section \ref{proposed_method}. The experimental results are presented in Section \ref{sec:exp}. Finally, Section \ref{sec:con} provides the conclusion.

\section{Related Work}\label{related}
In this section, we review graph-based HSI classification methods from two perspectives: pixel-level graph-based and superpixel-level graph-based methods. Based on recent research findings, we analyze the unresolved shortcomings of these two graph construction methods and subsequently highlight the advantages of our proposed approach.
\subsection{Pixel-level Graph-based Methods}
Under the motivation of applying graph to HSI classification, the most straightforward approach is to construct the graph using pixels as nodes, as shown in Fig.~\ref{motivation}(a). However, HSI contains a large number of pixels, \eg, in the classical HSI dataset Salinas, more than 50,000 pixels remain after removing the background. It presents significant challenges for training neural networks due to storage and computational difficulties associated with a large-scale graph matrix. As a result, research on pixel-level graph-based methods has been relatively limited in recent years, yet these findings offer important reference and guidance for future studies.

To perform HSI classification using a graph-based semi-supervised network, Mou~\etal~\cite {Nonlocal} proposed the nonlocal graph convolutional network to process the entire HSI, and used a non-local graph matrix to capture the relationship of all pixels. Considering the semantic gap between pixel-spectral information and high-level semantics, they performed graph learning instead of the fixed graph. In this process, constructing a fully connected graph for all pixels leads to increased computational overhead. To address the high computational cost caused by the graph matrix, Hong~\etal~\cite{miniGCN} introduced miniGCN. They constructed a random node sampler and applied it repeatedly to the entire graph until each node was sampled, generating a set of subgraphs. Each mini-batch processed a subgraph and applied graph convolution to it. Although this approach reduces computational complexity, the locality of the sampled subgraphs limits the full utilization of global spatial relationships. Su~\etal~\cite{graphcut} also constructed an undirected weighted graph for all pixels. They applied a graph-cut strategy to node embedding, which is used for the dimensionality reduction of HSI. Subsequently, they trained an extreme learning machine classifier using the generated low-dimensional data. In general, these pixel-level graph-based methods either incur high computational costs or sacrifice global information. However, our approach cleverly avoids both issues through a mini-batch training approach.
\subsection{Superpixel-level Graph-based Methods}
An increasing number of studies focus on constructing small graphs at the superpixel level for HSI~\cite{GIG,MView,ConGCN,MSSG,DMSGer,SGL,RLPA,SLGConv}, as illustrated in Fig.~\ref{motivation}(b). In these approaches, the image is first divided into multiple superpixels, and the features of the superpixel nodes are updated during training.

Earlier, Sellars~\etal~\cite{SGL} proposed a superpixel-level graph transductive learning framework for semi-supervised HSI classification. They also developed and implemented a novel modified state-of-the-art superpixel method specifically designed for HSI data, called hyper-manifold SLIC~\cite{SGL}. Yu~\etal~\cite{MView} proposed a graph-based multiview clustering, and SLIC~\cite{SLIC} is used to roughly divide HSI into many superpixels for avoiding the computational burden of graph Laplacian matrix. To acquire global and local features at the same time, Zhu~\etal~\cite{SLGConv} proposed short and long range graph convolution, and the raw HSI was first performed superpixel segmentation by SLIC~\cite{SLIC}. In order to apply graph contrastive learning to HSI classification, Yu~\etal~\cite{ConGCN} also utilized SLIC to obtain a set of compact superpixels. Besides, Jiang~\etal~\cite{MSSG} constructed a multi-scale spectral-spatial graph based on ERS~\cite{ERS}. Jiang~\etal~\cite{RLPA} used adaptive ERS~\cite{ERS} in their random label propagation.

Although computation cost is significantly reduced by the above superpixel-based methods, abundant information on the pixel level is inevitably ignored. The classification is performed at the superpixel level, which relies too much on the quality of image segmentation because inaccurate segmentation affects the performance of pixel-level classification.

Recently, some work has taken into account these limitations and supplemented by additional means to solve them. Yang~\etal~\cite{DMSGer} presented a new pixel-level graph nodes updating strategy to update pixel-level and superpixel-level features during the graph convolution at the same time. Jia~\etal~\cite{GIG} proposed a graph-in-graph model for HSI classification from a superpixel viewpoint, where the superpixel is not only the node of the external graph but also the internal pixel-level graph itself. These methods utilize pixel information as much as possible on the basis of superpixels and achieve some performance improvements, but pixel-level information, including pixel-spectral and pixel-spatial information, is still not fully utilized. In addition, even if the segmentation is very sufficient, it is impossible to ensure that pixels in a superpixel belong to the same class completely. The accuracy of superpixel segmentation is always a limitation for pixel-level classification.

For superpixel-level graph-based methods, naturally, the more thorough the segmentation, the better it can alleviate the aforementioned limitations. Ideally, the maximum number of image segments would equal the number of pixels. Thus, we favor constructing graphs at the pixel level, as shown in Fig.~\ref{motivation}(c), to fundamentally eliminate the constraints introduced by superpixels. In contrast to the previously mentioned pixel-level graph-based methods, we do not merely pursue pixel-by-pixel and mini-batch training approaches. Instead, we achieve this goal on the basis of leveraging global pixel information. Moreover, in order to make full use of label information resources and extract the supervised signals in unlabeled pixels, based on the lower bound estimation of mutual information, we propose the semi-supervised weighted contrastive loss to guide the learning of the encoder. Surprisingly, the loss can work well in the framework of mini-batch, which can be seen in the experiment section.
\section{Graph-Weighted Contrastive Learning}\label{proposed_method}
Given an HSI cube $\mathbf{H} \in \mathbb{R}^{M \times N \times\alpha}$, which consists of $\alpha$ number of spectral channels. Let $\h_i \in \mathbb{R}^{\alpha}, i\in\{1,\ldots,M \times N\}$ denote an HSI pixel. For the task of semi-supervised HSI classification, given a very small number of labeled pixels, unlabeled pixels are assigned to $c$ number of classes under the guidance of the proposed training target. The ground-truth label is a one-hot code, \eg, $\bm{t}_i=[1;0;\ldots;0]\in\{0,1\}^{c}$ means that the $i$-th pixel $\h_i$ belongs to the first class. The spatial coordinate of $\h$ is denoted by $(m,n)$, where $m\in\{1,\ldots,M\}$ and $n\in\{1,\ldots,N\}$. Let $G$ be a graph of pixels with vertex set $\mathbf{H}=[\h_1,\h_2,\ldots,\h_{M \times N}]$. The corresponding similarity graph is $S=[s_{ij}]$, where $s_{ij}$ is the pairwise similarity of pixels between $\h_i$ and $\h_j$. We train a neural network, $f(\h_i, \h_j | \theta)$, where $\theta$ represents the model parameters. The network is designed to process pair-wise pixel information, feeding it pairs of pixel features $\h_i$ and $\h_j$ to contrast the relationships between pair-wise pixels.
\subsection{Contrastive Learning}
As a widely adopted self-supervised learning technique, contrastive learning aims to capture meaningful features by maximizing the mutual information between two different views of the data~\cite{CPC,MINE,DSSN2023,infomatch}. Mutual information measures the degree of dependence between two random variables, indicating how much the uncertainty of one variable is reduced by knowing information about the other. In the context of unsupervised feature learning, the goal is to extract meaningful features from the dataset $\mathcal{X}=\{\x_i\}_{i=1}^{M\times N}$. Consequently, contrastive learning can be interpreted as an approach to maximize mutual information, leading to the following optimization objective:
\begin{equation}
	\max {\rm I}(\mathcal{X}^{(1)} ; \mathcal{X}^{(2)})
	=\max {\rm div}_{\rm kl}
	\big(p(\x^{(1)}, \x^{(2)}) \|
	p(\x^{(1)}) p(\x^{(2)})\big)
	\label{mi}
\end{equation}
where $\mathcal{X}^{(1)}$ and $\mathcal{X}^{(2)}$ are two views of $\mathcal{X}$, ${\rm div}_{\rm kl}$ denotes the Kullback-Leibler divergence, $p(\x^{(1)},\x^{(2)})$ is the joint probability of the pairwise views, and $p(\x^{(1)})p(\x^{(2)})$ denotes the product of their marginal probabilities.

Although mutual information plays a critical role, directly calculating it is often challenging, particularly for high-dimensional data. To address this issue, researchers have developed various estimation methods~\cite{IBA,Nguyen,fgan,CPC,MINE,DSSN2023,infomatch,MILBO} based on lower bounds. Although these methods cannot directly compute the true value of mutual information, they offer a feasible optimization objective for maximizing mutual information on high-dimensional data. Following contrastive learning~\cite{CPC,MINE,DSSN2023,infomatch}, Jensen-Shannon divergence is employed as a replacement for Kullback-Leibler divergence~\cite{fgan,DSSN2023,infomatch}, and the lower bound is estimated to be
\begin{align}
	&
	{\rm div}_{\rm js}
	\big(p(\x^{(1)}, \x^{(2)}) \| p(\x^{(1)}) p(\x^{(2)})\big) \geqslant \notag\\
	&  \mathbb{E}_{p(\x^{(1)}, \x^{(2)})}
	\log \big(f(\z^{(1)}, \z^{(2)} | \theta)\big)\notag\\
	+& \mathbb{E}_{p(\x^{(1)}) p(\x^{(2)})}
	\log \big(1-f(\z^{(1)}, \z^{(2)} | \theta)\big)
\end{align}
where $\mathbb{E}_{p(\x^{(1)}, \x^{(2)})}\log \big(f(\z^{(1)}, \z^{(2)} | \theta)\big)$ denotes the expectation of $\log f(\cdot,\cdot)$ with respect to $p(\x^{(1)}, \x^{(2)})$, given that $\x$ and $\z$ form a one-to-one input-output pair. Here, $f(\cdot,\cdot)$ is the similarity score of pairs, $\z^{(1)}$ and $\z^{(2)}$ are latent features generated by an encoder.

Maximizing this lower bound is equivalent to minimizing the following contrastive loss~\cite{DSSN2023,infomatch}
\begin{align}
	\mathcal{L}_{\textrm{contrastive}}
	= & -\frac{1}{|\mathcal{P}|} \sum_{(i, i) \in \mathcal{P}} \log f(\z_i^{(1)}, \z_i^{(2)} | \theta)\notag\\
	&-\frac1{|\mathcal{N}|} \sum_{(i, j) \in \mathcal{N}}
	\log \big(1-f(\z_i^{(1)}, \z_j^{(2)} | \theta)\big)\label{lcontrastive}
\end{align}
where $\mathcal{P}$ denotes the set of positive samples, and $\mathcal{N}$ denotes the set of negative samples. This loss illustrates the classic concept of contrastive learning, \ie, shortening the distance between similar samples and keeping different samples away from each other.

Inspired by BYOL~\cite{BYOL}, many contrastive learning algorithms only use the positive pairs~\cite{DSSN2023,infomatch}. The similarity $f(\cdot,\cdot)$ of positive pairwise is given by~\cite{DSSN2023,infomatch},
\begin{align}\label{eq-gaussian}
	f(\z^{(1)}_i,\z_i^{(2)}|\theta)=\exp\Bigl(-{\bigl\|\z^{(1)}_i-\z_i^{(2)}\bigr\|^2_2}\Bigr)\,,
\end{align}
which indicates that before applying the contrastive loss, it is essential to construct positive pairs from different views. Moreover, the effectiveness of contrastive learning largely depends on the quality of these constructed positive pairs. 
\subsection{Graph-weighted Contrastive Learning}
We assume that each positive pair is connected by an edge, and the pair-wise relations are represented by an indicator matrix $A=[a_{ij}]$,
\begin{align}
	a_{ij}=
	\begin{cases}
		1, &{\rm if}\, s_{ij}>0\,,\\
		0, &{\rm otherwise}\,.
	\end{cases}
\end{align}

Positive pairs are constructed based on the presence of an edge, and a pair-wise similarity is defined by
\begin{align}\label{wg-gaussian}
	f(\z_i,\z_j|\theta)=\exp\Bigl(-{a_{ij}\bigl\|\z_i-\z_j\bigr\|^2_2}\Bigr)\,.
\end{align}
The element $a_{ij}$ in the adjacency matrix $A$ indicates the presence of an edge between two nodes $i$ and $j$, whereas the element $s_{ij}$ in the similarity matrix $S$ represent the probability of an edge existing between them. Given the predefined similarity matrix $S=[s_{ij}]$, we interpret the similarity as the probability that two elements belong to positive pairs. Based on this observation, we define the graph-weighted similarity by
\begin{align}\label{wg-gaussian_}
	f(\z_i,\z_j|\theta)=\exp\Bigl(-{ s_{ij}\bigl\|\z_i-\z_j\bigr\|^2_2}\Bigr)\,.
\end{align}
Substituting Eq.~\eqref{wg-gaussian_} into Eq.~\eqref{lcontrastive}, we have the graph-weighted contrastive loss function,
\begin{align}
	\mathcal{L}_{\textrm{gwcl}}
	=  \frac{1}{|\mathcal{P}|} \sum_{(i, j) \in \mathcal{P}}s_{ij}\|\z_i-\z_j\bigr\|^2_2\,.\label{gwcl}
\end{align}

We use $\mathcal{L}_{\textrm{gwcl}}$, Eq.~\eqref{gwcl}, for the labeled and unlabeled pixels in the mini-batch training.

Additionally, to fully utilize the label pixels, the cross-entropy loss function between the labels and its prediction is given by
\begin{equation}
	\mathcal{L}_{\rm ce}
	=- \sum_{i=1}^{l} \sum_{k=1}^c t_{i k} \log z_{ik}\label{loss_ce}
\end{equation}
where \(l\) denotes the number of labeled samples, \(c\) is the number of classes, and \(t_{ik}\) is the \(k\)-th element of the ground-truth label \(\bm{t}_i\) for the \(i\)-th sample.

To summarize, the overall loss function is given by
\begin{equation}
	\mathcal{L}
	=\mathcal{L}_{\rm gwcl}+\lambda\mathcal{L}_{\rm ce}
	\label{Loss}
\end{equation}
where $\lambda$ is used to balance the contributions of the two loss terms: the graph-based weighted contrastive loss \(\mathcal{L}_{\rm gwcl}\) and the cross-entropy loss \(\mathcal{L}_{\rm ce}\).
\subsection{GWCL Algorithm}
\subsubsection{Graph Construction}
At first, we reduce the dimension of HSI from $\h\in\mathbb R^\alpha$ to $\h'\in\mathbb R^\beta$ with the ARM algorithm~\cite{ARM2017}, and the pixel-wise feature $\x$ is defined by $\x=[\h';m;n]$.
The integration of pixel-spectral and pixel-spatial information is achieved by concatenating pixel-spectral features with pixel-spatial coordinates.

Then, we define the graph similarity by
\begin{equation}
	s_{ij}=
	\begin{cases}
		\exp\left\{
		-\frac12(\x_i-\x_j)^\top\Sigma^{-1}(\x_i-\x_j)
		\right\}, &{\rm if}\, \x_i\in K\textrm{-NN}(\x_j)\\
		0, &{\rm otherwise}
	\end{cases}
	\label{sij}
\end{equation}
where \(\Sigma={\rm diag}([\bm{1}_{\beta};\sigma_{m};\sigma_n])\) is a diagonal matrix that controls the feature-wise scaling. Specifically, \(\bm{1}_{\beta}\) represents a vector of ones with length \(\beta\), ensuring equal weighting for the first \(\beta\) dimensions, while \(\sigma_m\) and \(\sigma_n\) are hyperparameters that balance the influence of the remaining feature dimensions. The choice of \(K\)-nearest neighbors ($K$-NN) based on the Mahalanobis distance is motivated by its ability to account for correlations between features and scale differences, unlike the Euclidean distance, which assumes isotropic feature distribution. By using Mahalanobis distance, we improve the robustness of neighbor selection, especially in high-dimensional or heterogeneous feature spaces, leading to a more meaningful similarity structure in the graph.

\subsubsection{Mini-batch Training}
Our objective is to integrate the proposed loss function into the mini-batch training framework. The graph $S$ is pre-constructed, and during training, all labeled pixels along with a mini-batch of unlabeled pixels are simultaneously loaded. The pair-wise similarities between pixels in the current mini-batch are extracted from graph $S$, allowing the model to leverage global similarity information while addressing computational challenges.

In the mini-batch training process, not all graph nodes need to be available at once. Each mini-batch includes only a subset of graph nodes for gradient computation and parameter updates, which demonstrates GWCL's ability to generalize to unseen nodes. To further improving the performance, the model undergoes a pre-training stage where it is trained using all labeled data. To sum up, our GWCL approch for semi-supervised HSI classification algorithm is shown in Algorithm~\ref{algo}. The network $f(\h_i, \h_j | \theta)$ is a two-layer fully connected multilayer perceptron with 180 nodes in the hidden layer, followed by a softmax layer at the output logit.
\begin{algorithm}[!t]
	\caption{GWCL for semi-supervised HSI classification.}\label{algo}
	\begin{algorithmic}[1]
		\Require $\mathcal{D}_l = \{(\h_1,\bm{t}_1),\ldots,(\h_l,\bm{t}_l)\}$ and $\mathcal{D}_u = \{\h_{l+1},\ldots,\h_{l+M\times N}\}$.
		\Ensure Hyperparameters $\lambda$, $\sigma_m$, and $\sigma_n$. The mini-batch size $b$ and the number of the reduced dimension $\beta$. For the model $f(\cdot,\cdot|\theta)$, we initialize its parameter $\theta$.
		\State Reduce dimension of pixels $\h'\in\mathbb R^{\beta}\leftarrow\h\in\mathbb R^{\alpha}$ by ARM~\cite{ARM2017}\,;
		\State Integrate pixel-spectral and pixel-spatial features by $\x=[\h',m,n]$\,;
		\State Construct the graph by Eq.~\eqref{sij}\,.
		\State Pre-train $f(\cdot,\cdot|\bm{\theta})$ with $\mathcal{D}_l$ and the supervised loss Eq.~\eqref{loss_ce}.
		\State Train model $f(\cdot,\cdot|\bm{\theta})$ in mini-batch way with the whole pixels and the overall loss Eq.~\eqref{Loss}\,.
		\Return The optimized parameter $\theta^\star$.
	\end{algorithmic}
\end{algorithm}
\section{Experiments}\label{sec:exp}
\subsection{Experimental Setting}
Three widely used HSI datasets, \ie, Indian Pines, Salinas, and University of Pavia are adopted to demonstrate the effectiveness of GWCL. A brief introduction to these datasets is provided below.
\begin{figure}[!t]
	\centering
	\includegraphics[width=0.48\textwidth]{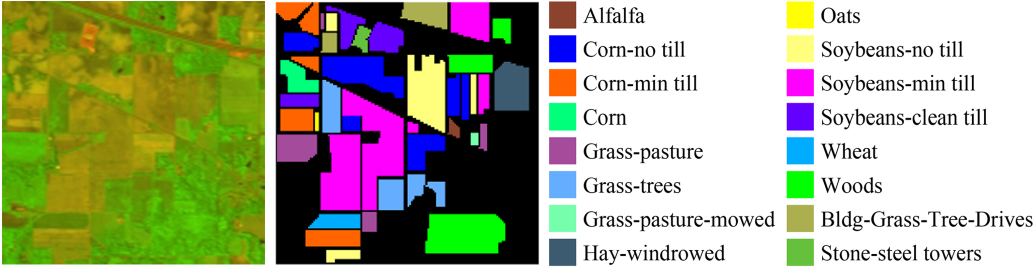}
	\caption{Indian Pines in pseudocolor and the ground-truth map with class information.}\label{ipdata}
\end{figure}

Indian Pines was taken in Indiana by an AVIRIS sensor. The data size is 145$\times$145, with 224 channels, of which 200 are effective and span the spectral range of $[400,2500]$nm. The data has a total of 16 land classes and a spatial resolution of 20m. The image in pseudocolor and ground-truth map with class information are shown in Fig.~\ref{ipdata}.

\begin{figure}[!t]
	\centering\includegraphics[width=0.48\textwidth]{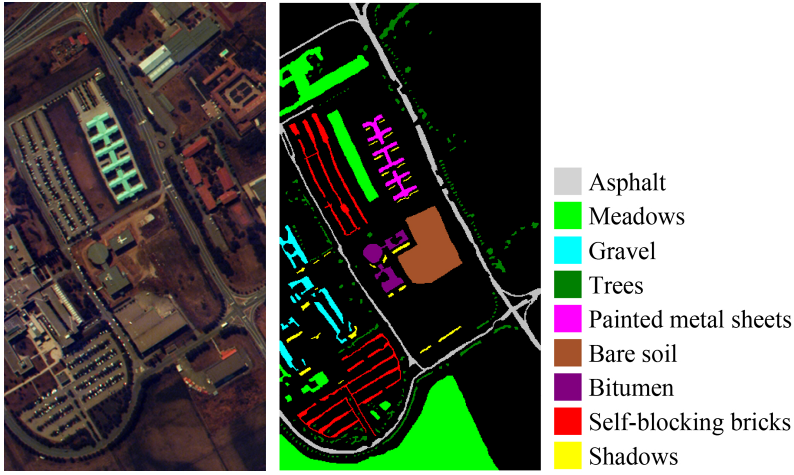}\\
	\caption{Salinas in pseudocolor and ground-truth map with class information.}\label{sadata}
\end{figure}

Salinas was photographed by AVIRIS sensor in Salinas Valley, California. The spatial resolution of the data is 3.7m and the size is 512$\times$217. There are 224 channels in the original data, and 204 channels in the spectral range of $[360,2500]$nm are left after removing the channels with serious water vapor absorption. The data contains 16 land classes. The image in pseudocolor and ground-truth map with class information are shown in Fig.~\ref{sadata}.

\begin{figure}[!t]
	\centering
	\includegraphics[width=0.48\textwidth]{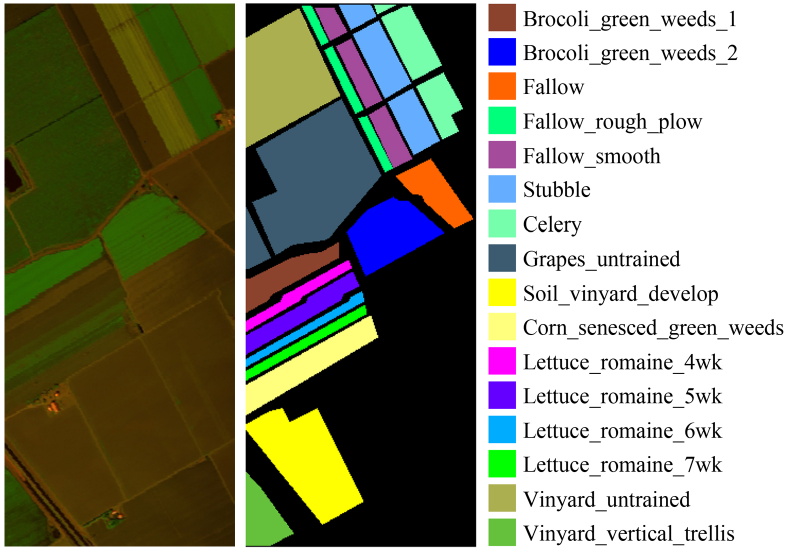}\\
	\caption{University of Pavia in pseudocolor and ground-truth map with class information.}\label{pudata}
\end{figure}

University of Pavia was acquired by ROSIS sensor. The sensor has a total of 115 channels, and there are 103 channels in the spectral range of $[430,860]$nm left after processing. In addition, it contains nine land classes with a spatial resolution of 1.3m and a size of 610$\times$340. The image in pseudocolor and ground-truth map with class information are shown in Fig.~\ref{pudata}.

For each dataset, 30 labeled pixels are randomly selected from each land cover class for training. If a class contains fewer than 30 pixels, 15 labeled pixels are chosen instead. The details regarding the volumes of the training and testing sets for different datasets can be found in Tables~\ref{ipclass}, \ref{saclass}, and \ref{puclass}. This approach is consistent with the settings used in other baseline methods.

\begin{table}[t]
	\caption{Number of pixels within the training and testing sets of all classes in Indian Pines.}
	\label{ipclass}
	\centering
	\resizebox{0.8\linewidth}{!}{ %
		\begin{tabular}{c l|c r}
			\toprule
			\multicolumn{2}{c|}{\textmd{Class}} & \multicolumn{2}{c}{\textmd{Number}} \\
			\midrule
			$\textmd{Code}$&$\textmd{Name}$&$\textmd{Train}$&$\textmd{Test}$\\
			\midrule
			1 &Alfalfa &30 &16\\
			2 &Corn-no till &30 &1398\\
			3 &Corn-min till &30  &800\\
			4 &Corn &30 &207\\
			5 &Grass-pasture &30 &453\\
			6 &Grass-trees &30 &700\\
			7 &Grass-pasture-mowed &15 &13\\
			8 &Hay-windrowed &30  &448\\
			9 &Oats &15 &5\\
			10 &Soybeans-no till &30   &942\\
			11 &Soybeans-min till &30 &2425\\
			12 &Soybeans-clean till &30  &563\\
			13 &Wheat &30  &175\\
			14 &Woods &30 &1235\\
			15 &Bldg-Grass-Tree-Drives &30  &356\\
			16 &Stone-steel towers &30  &63\\
			\bottomrule
	\end{tabular} }
\end{table}

\begin{table}[t]
	\caption{Number of pixels within the training and testing sets of all classes in Salinas.}
	\label{saclass}
	\centering
	\resizebox{0.8\linewidth}{!}{ %
		\begin{tabular}{c l|c r}
			\toprule
			\multicolumn{2}{c|}{\textmd{Class}} & \multicolumn{2}{c}{\textmd{Number}} \\
			\midrule
			$\textmd{Code}$&$\textmd{Name}$&$\textmd{Train}$&$\textmd{Test}$\\
			\midrule
			1 &Brocoli\_green\_weeds\_1 &30 &1979\\
			2 &Brocoli\_green\_weeds\_2 &30 &3696\\
			3 &Fallow &30  &1946\\
			4 &Fallow\_rough\_plow &30 &1364\\
			5 &Fallow\_smooth &30 &2648\\
			6 &Stubble &30 &3929\\
			7 &Celery &30 &3549\\
			8 &Grapes\_untrained &30  &11241\\
			9 &Soil\_vinyard\_develop &30 &6173\\
			10 &Corn\_senesced\_green\_weeds &30   &3248\\
			11 &Lettuce\_romaine\_4wk &30 &1038\\
			12 &Lettuce\_romaine\_5wk &30  &1897\\
			13 &Lettuce\_romaine\_6wk &30  &886\\
			14 &Lettuce\_romaine\_7wk &30 &1040\\
			15 &Vinyard\_untrained &30  &7239\\
			16 &Vinyard\_vertical\_trellis &30  &1777\\
			\bottomrule
	\end{tabular} }
\end{table}

\begin{table}[t]
	\caption{Number of pixels within the training and testing sets of all classes in University of Pavia.}\label{puclass}
	\centering
	\resizebox{0.8\linewidth}{!}{ %
		\begin{tabular}{c l|c r}
			\toprule
			\multicolumn{2}{c|}{\textmd{Class}} & \multicolumn{2}{c}{\textmd{Number}} \\
			\midrule
			$\textmd{Code}$&$\textmd{Name}$&$\textmd{Train}$&$\textmd{Test}$\\
			\midrule
			1 &Asphalt &30 &6601\\
			2 &Meadows &30 &18619\\
			3 &Gravel &30  &2069\\
			4 &Trees &30 &3034\\
			5 &Painted metal sheets &30 &1315\\
			6 &Bare soil &30 &4999\\
			7 &Bitumen &30 &1300\\
			8 &Self-blocking bricks &30  &3652\\
			9 &Shadows &30 &917\\
			\bottomrule
	\end{tabular} }
\end{table}

According to the majority of related studies, three indices are used to quantitatively evaluate classification performance in this paper: Overall Accuracy (OA), Average Accuracy (AA)\cite{foody2002status}, and the $\kappa$ coefficient\cite{cohen1960coefficient}.

OA measures the model's general performance across all classes and is given by
\begin{align}
	\text{OA} = \frac{\text{Number of Correct Predictions}}{\text{Total Number of Predictions}}   \,.
\end{align}
A high OA indicates strong overall performance but can be misleading in imbalanced datasets where one class dominates.

AA addresses the limitations of OA by considering each class individually, making it particularly useful when dealing with class imbalance. AA is the average of per-class recall and is given by
\begin{align}
	\text{AA} = \frac{1}{c} \sum_{k=1}^{c} \frac{\text{Correctly Classified Samples in Class } k}{\text{Total Samples in Class } k}\,.
\end{align}

$\kappa$ assesses the agreement between the observed classification and a classification made by chance, adjusting for random chance agreement.
\begin{align}
	\kappa = \frac{\text{OA} - P_e}{1 - P_e}
\end{align}
where \(P_e\) is the expected agreement by chance, calculated based on the distribution of class labels. It evaluates how well the model predictions align with the true labels beyond random guessing.

The pixel-spectral feature dimension after dimensionality reduction is set to $\beta=20$. In the pre-training stage, the number of epoch is set to 300, and the mini-batch size is set to 1. In the mini-batch training stage, the number of epoch is set to 1000, and the mini-batch size is set to 512. The trade-off coefficient $\lambda$ in semi-supervised contrastive loss is set to 8. In addition, the position coordinates of the pixels are min-max normalized. Other parameters are shown in Table~\ref{parameters} where $\eta_1$ and $\eta_2$ are the learning rates of the two training stages, respectively. Each experiment is repeated ten times and the mean and standard deviation are recorded.

\begin{table}[!t]
	\caption{Hyperparameters for different datasets.
	}
	\label{parameters}
	\centering
	\scalebox{1.0}{
		\begin{tabular}{l| l l l l l}
			\toprule
			{Parameters}  &$\eta_1$ &$\eta_2$  &$\sigma_{m}$ &$\sigma_{n}$ &$K$\\
			\midrule
			Indian Pines  &0.001   &0.001 &0.04 &0.001 &10\\
			Salinas  &0.001   &0.001 &0.04 &0.04 &10  \\
			University of Pavia  &0.005   &0.01 &1 &0.4 &50 \\
			\bottomrule
	\end{tabular}}
\end{table}
\subsection{Comparison}
To evaluate the performance of our approach, GWCL is compared with several state-of-the-art semi-supervised HSI classification methods. First, there are two traditional HSI classification methods, including image fusion and recursive filtering (IFRF)~\cite{IFRF17} and albedo recovery for HSI classification (ARM)~\cite{ARM2017}. In particular, four superpixel-level graph-based methods are employed for comparison, including superpixel graph learning (SGL)~\cite{SGL}, multilevel superpixel structured graph U-net (MSSGU)~\cite{MSSGU}, multiscale dynamic GCN (DMSGer)~\cite{DMSGer} and a GCN model with contrastive learning (ConGCN)~\cite{ConGCN}.

\begin{figure}[!t]
	\centering
	\includegraphics[width=0.48\textwidth]{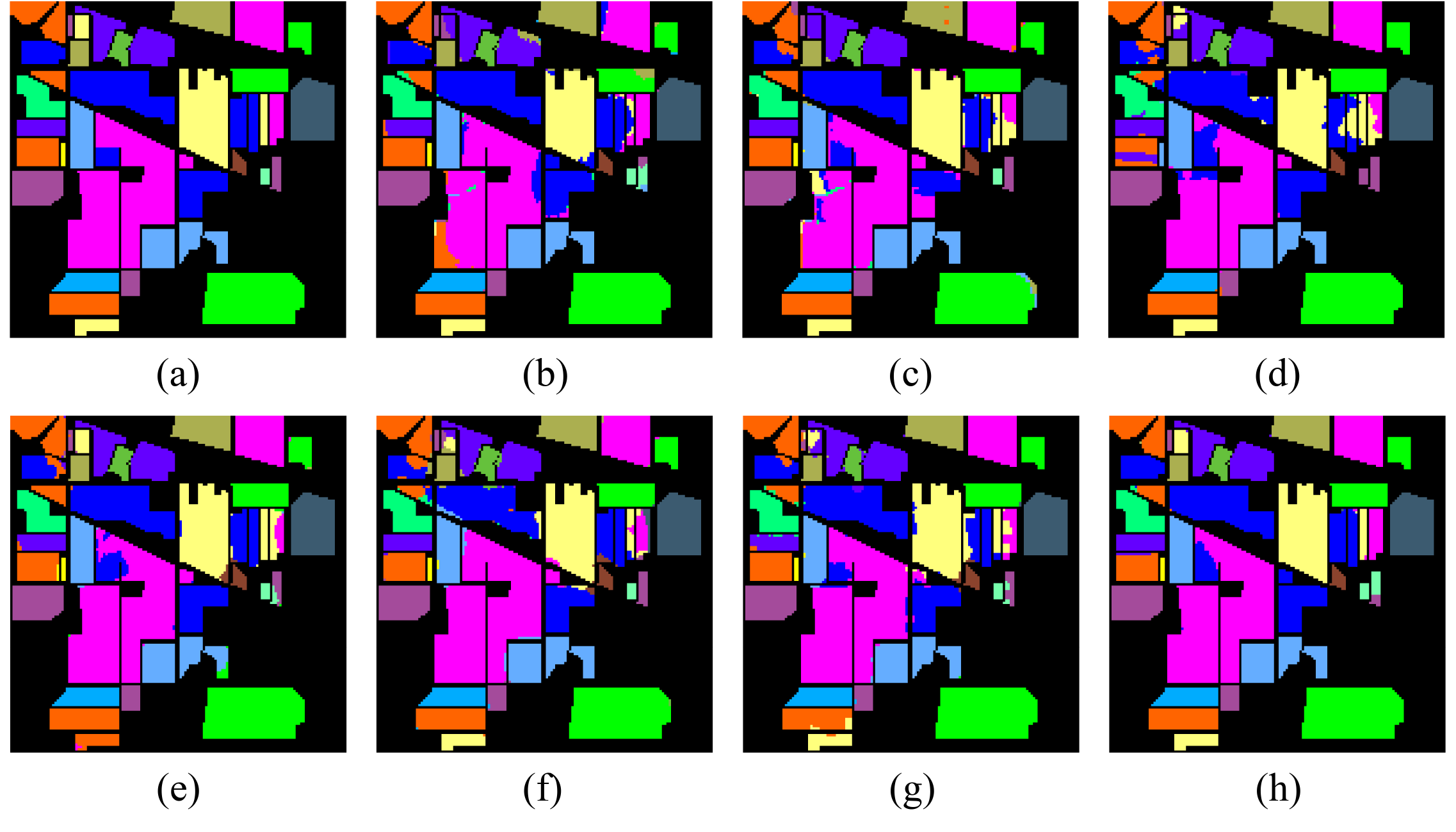}\\
	\caption{The ground-truth and classified maps of different methods on the Indian Pines dataset. (a) the ground-truth map; (b) IFRF ($92.65\%$); (c) ARM ($94.18\%$); (d) SGL ($94.31\%$); (e) MSSGU ($95.93\%$); (f) DMSGer ($95.25\%$); (g) ConGCN ($96.44\%$); (h) GWCL ($98.81\%$) .}\label{ipmap}
\end{figure}

\begin{figure}[!t]
	\centering
	\includegraphics[width=0.48\textwidth]{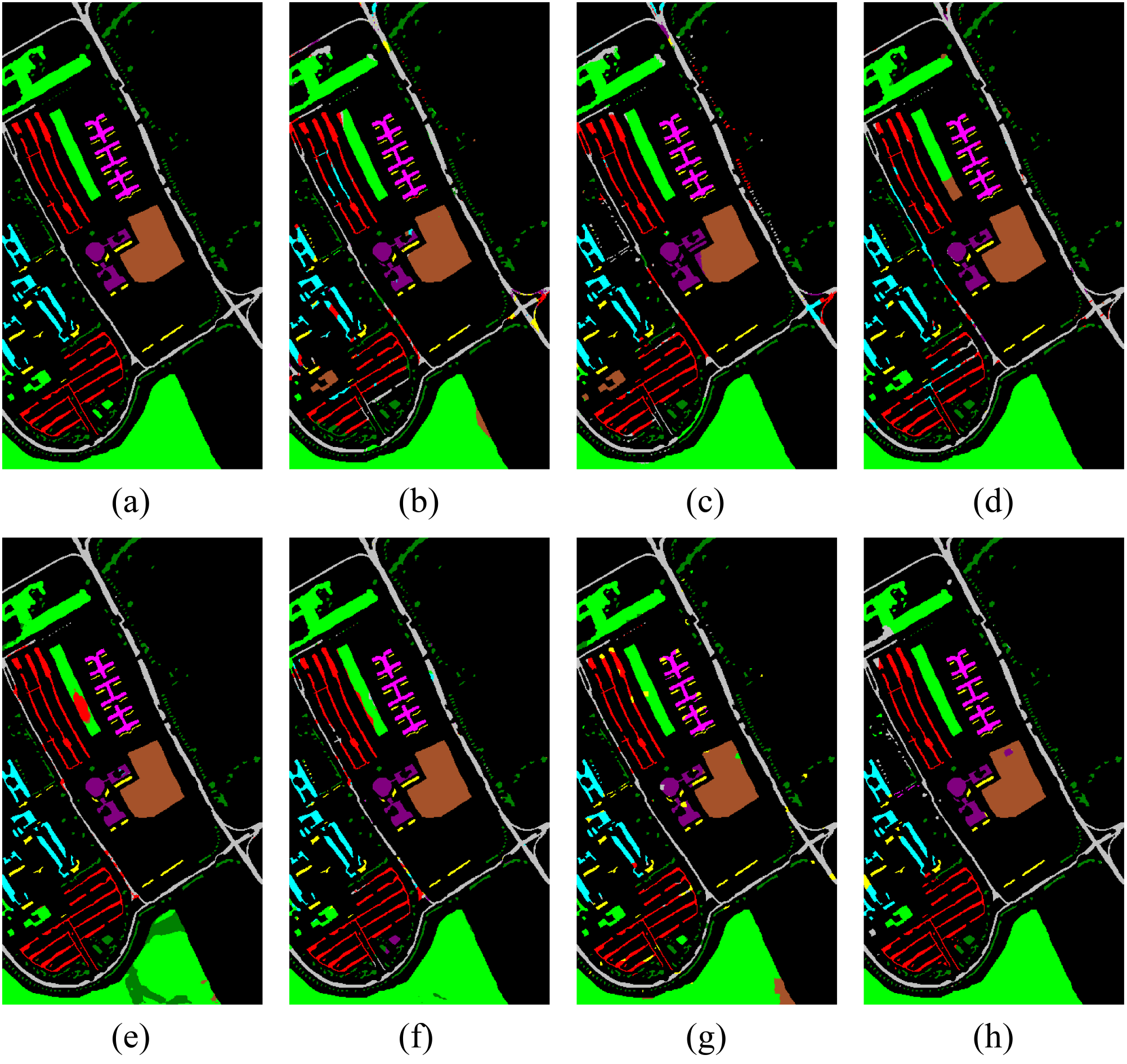}
	\caption{The ground-truth and classified maps of different methods on Salinas. (a) the ground-truth map; (b) IFRF ($98.53\%$); (c) ARM ($99.17\%$); (d) SGL ($99.01\%$); (e) MSSGU ($99.48\%$); (f) DMSGer ($99.63\%$); (g) ConGCN ($99.08\%$); (h) GWCL ($99.96\%$) .}\label{samap}
\end{figure}

\begin{figure}[!t]
	\centering
	\includegraphics[width=0.48\textwidth]{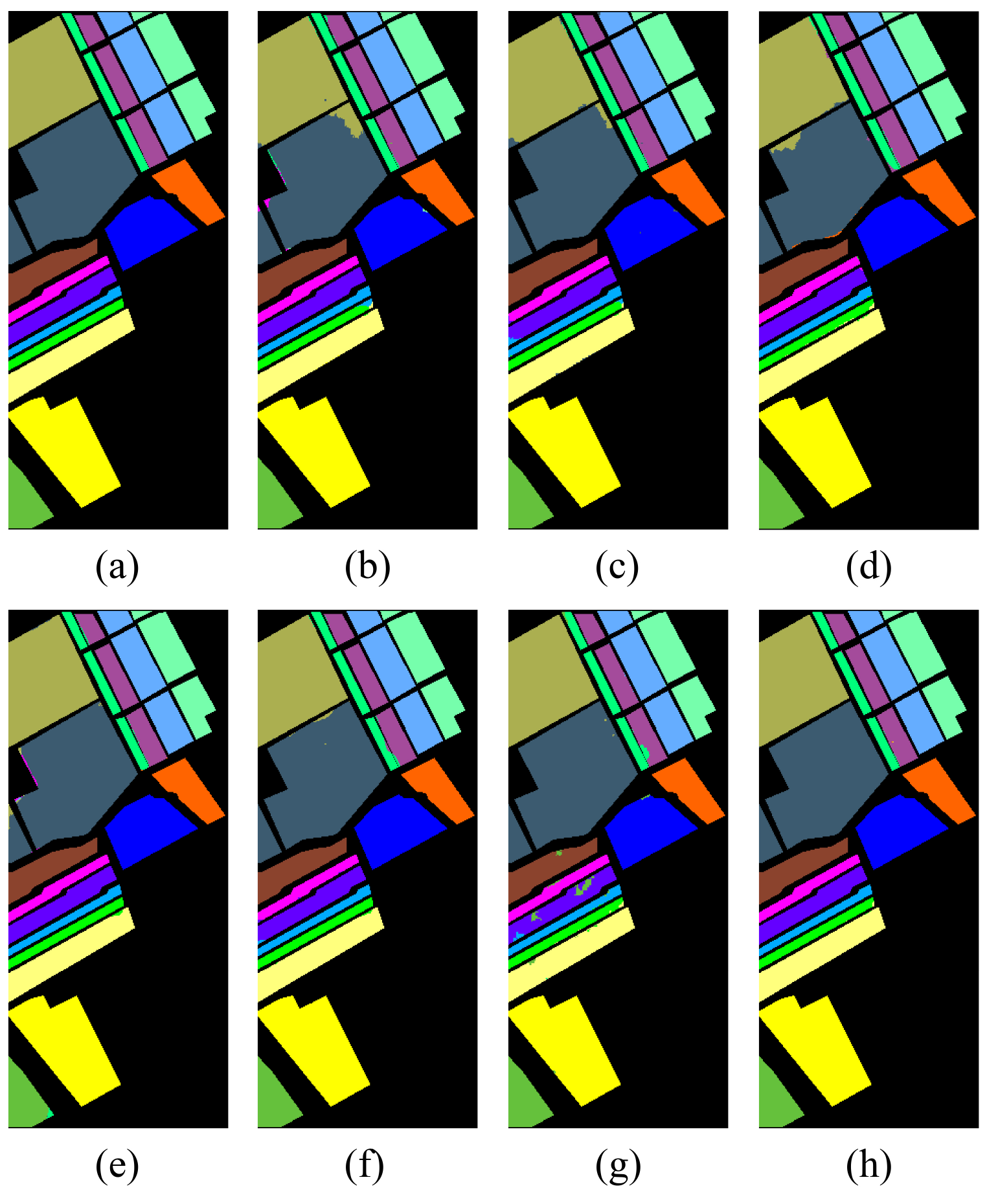}
	\caption{The ground-truth and classified maps of different methods on University of Pavia. (a) the ground-truth map; (b) IFRF ($90.32\%$); (c) ARM ($90.76\%$); (d) SGL ($95.07\%$); (e) MSSGU ($91.62\%$); (f) DMSGer ($96.66\%$); (g) ConGCN ($95.27\%$); (h) GWCL ($97.46\%$) .}\label{pumap}
\end{figure}

Tables~\ref{accuracy1}, \ref{accuracy2}, and \ref{accuracy3}
present the classification performance of different methods on the Indian Pines, Salinas, and University of Pavia datasets, respectively. The visualized results as shown in Figures~\ref{ipmap}, \ref{samap}, and \ref{pumap}. They include the classification performance for each class, along with three overall performance metrics, expressed as percentages.
The standard deviation is provided in parentheses. The best and second-best results for the three overall performance metrics are highlighted in bold and underlined, respectively.

\begin{table*}[t]
	\caption{Classification performance on the Indian Pines dataset. The highest value is bolded, and the second-best is underlined.}
	\label{accuracy1}
	\vspace{1mm}
	\centering
	\resizebox{1.0\linewidth}{!}{ %
		\begin{tabular}{l|rrrrrr|r}
			\toprule $\textmd{Class}$&$\textmd{IFRF\cite{IFRF17}}$&$\textmd{ARM\cite{ARM2017}}$&$\textmd{SGL\cite{SGL}}$&$\textmd{MSSGU\cite{MSSGU}}$&$\textmd{DMSGer\cite{DMSGer}}$&$\textmd{ConGCN\cite{ConGCN}}$&$\textmd{GWCL}$\\
			\midrule
			1 &77.49(35.90) &99.41(1.86) &100.00(0.00) &100.00(0.00) &100.00(0.00) &98.75(3.75) &100.00(0.00)\\
			2 &93.25(3.40) &89.26(4.80) &89.08(4.56) &91.69(1.52)  &90.13(0.85) &92.07(1.55) &97.19(2.38)\\
			3 &84.41(9.04) &88.87(6.31) &90.00(3.24)  &98.35(0.60)  &95.45(1.19) &97.50(0.64) &99.77(0.45)\\
			4 &84.68(9.64) &91.71(3.63) &97.10(4.43) &98.13(0.56) &100.00(0.00)  &100.00(0.00) &100.00(0.00)\\
			5 &94.42(6.21) &99.27(0.69) &97.75(1.73) &95.92(0.81)  &97.39(0.43) &94.50(1.79) &97.84(2.26)\\
			6 &96.85(2.18) &98.64(1.09) &99.30(0.42) &99.84(0.23)  &97.99(0.39) &98.99(0.04) &100.00(0.00)\\
			7 &62.99(29.26) &71.48(30.53) &100.00(0.00) &100.00(0.00) &100.00(0.00)  &100.00(0.00) &100.00(0.00)\\
			8 &100.00(0.00) &99.73(0.46) &100.00(0.00)  &100.00(0.00) &100.00(0.00) &100.00(0.00) &100.00(0.00)\\
			9 &54.22(23.22) &71.02(32.25) &100.00(0.00) &100.00(0.00) &100.00(0.00)  &100.00(0.00) &100.00(0.00)\\
			10 &87.84(5.35) &89.53(4.78) &90.50(6.12)   &96.26(1.52)  &86.31(1.42)  &93.57(2.86) &96.66(4.39)\\
			11 &95.53(2.00) &97.50(1.91) &94.76(3.50) &91.54(0.78) &95.43(0.23)  &97.12(1.19) &98.78(1.63)\\
			12 &89.44(5.78) &93.20(5.42) &94.33(2.67)  &98.47(0.51) &97.98(0.70) &97.57(0.87) &98.63(1.29)\\
			13 &91.99(4.87) &98.24(2.01) &99.09(0.38)  &100.00(0.00) &100.00(0.00) &100.00(0.00) &99.83(0.39)\\
			14 &99.30(0.63) &99.19(0.72) &99.84(0.24) &99.98(0.04) &99.98(0.03)  &99.83(0.02) &100.00(0.00)\\
			15 &92.33(4.26) &89.54(10.07) &99.58(0.42)  &99.77(0.36) &100.00(0.00) &99.41(0.08) &99.44(0.75)\\
			16 &96.59(1.32) &98.55(0.51) &100.00(0.00)  &100.00(0.00) &99.68(0.64) &87.14(2.18) &98.89(0.77)\\
			\midrule
			OA (\%) &92.56(1.55) &94.12(1.28) &94.35(0.93)  &95.87(0.11) &95.39(0.21) &\underline{96.74(0.50)}  &\textbf{98.81(0.33)}\\
			AA (\%) &87.58(3.02) &92.20(3.79) &84.46(0.42)   &\underline{98.12(0.09)} &97.52(0.12) &97.28(0.29) &\textbf{99.19(0.25)}\\
			$\kappa$ &91.51(1.74) &93.29(1.44) &93.53(1.06)  &95.28(0.13) &94.72(0.23) &\underline{96.27(0.57)} &\textbf{98.56(0.43)}\\
			\bottomrule
	\end{tabular} }
\end{table*}

(1) \textit{Results on Indian Pines}: Overall, as shown in Table~\ref{accuracy1}, GWCL achieves the highest OA, AA and $\kappa$ coefficient and outperforms the second best records by $2.07\%$, $1.07\%$ and $2.29\%$ respectively. Besides, the standard deviations of GWCL are relatively small, which means the classification results by GWCL are more stable. In terms of specific classes, GWCL, like DMSGer, achieves $100\%$ accuracy in seven classes. However, GWCL performs well in all classes, and achieves an accuracy of $96.66\%$ in the worst-performing Class 10. In contrast, other methods have classes with relatively poor performance, \eg, the accuracy of DMSGer in Class 10 is only $86.31\%$, and the accuracy of ConGCN in Class 16 is only $87.14\%$. It may indicate that GWCL in Indian Pines is less affected by the location and shape of various land covers. The visualization of the classification results of various methods is shown in Fig.~\ref{ipmap}. For the sake of clarity, ground-truth map of Indian Pines is first exhibited in (a), and the OA values of different methods are provided. In contrast, the classification mapping of GWCL is significantly closer to ground-truth, and other methods are highly confusing in the upper left corner and the upper right part.

\begin{table*}[t]
	\caption{Classification performance on the Salinas dataset. The highest value is bolded and the second-best is underlined.}
	\label{accuracy2}
	\centering
	\resizebox{1.0\linewidth}{!}{ %
		\begin{tabular}{l|rrrrrr|r}
			\toprule        $\textmd{Class}$&$\textmd{IFRF\cite{IFRF17}}$&$\textmd{ARM\cite{ARM2017}}$&$\textmd{SGL\cite{SGL}}$&$\textmd{MSSGU\cite{MSSGU}}$&$\textmd{DMSGer\cite{DMSGer}}$&$\textmd{ConGCN\cite{ConGCN}}$&$\textmd{GWCL}$\\
			\midrule
			1&99.98(0.06) &100.00(0.00) &100.00(0.00) &100.00(0.00) &98.54(0.51) &100.00(0.00) &100.00(0.00)\\
			2 &100.00(0.00) &100.00(0.00) &100.00(0.00)  &99.99(0.03) &99.80(0.30) &100.00(0.00) &100.00(0.01)\\
			3 &99.79(0.26) &99.85(0.02) &100.00(0.00)  &100.00(0.00) &99.98(0.04)  &100.00(0.00) &100.00(0.00)\\
			4 &96.89(1.44) &97.52(0.83) &98.15(0.85) &99.88(0.07) &98.96(0.37) &98.50(0.82) &99.76(0.51)\\
			5 &99.68(0.44) &99.48(0.84) &98.36(0.05) &99.50(0.17) &99.85(0.10) &97.58(0.60) &99.67(0.09)\\
			6 &99.99(0.02) &100.00(0.00) &100.00(0.00) &100.00(0.00) &100.00(0.00) &99.81(0.07) &99.99(0.01)\\
			7 &99.66(0.16) &99.77(0.09) &99.89(0.12) &100.00(0.00) &99.90(0.09) &99.94(0.01) &100.00(0.00)\\
			8 &99.06(1.39) &99.39(0.88) &98.52(0.39) &97.65(0.70) &99.80(0.09) &98.33(1.16) &99.92(0.06)\\
			9 &99.95(0.13) &100.00(0.01) &100.00(0.00)  &100.00(0.00) &100.00(0.00) &100.00(0.00) &100.00(0.01)\\
			10 &99.53(0.62) &99.73(0.01) &97.98(1.70) &99.47(0.10) &98.99(0.26) &99.28(0.68) &99.85(0.09)\\
			11 &99.17(0.30) &99.88(0.27) &97.86(2.27) &100.00(0.00) &95.88(1.60)  &99.74(0.09) &100.00(0.00)\\
			12 &99.66(0.36) &99.85(0.26) &99.74(0.60) &99.87(0.17) &99.77(0.29) &98.26(0.71) &100.00(0.00)\\
			13 &97.78(1.40) &97.76(2.88) &98.70(0.06) &100.00(0.00) &99.66(0.17) &97.58(0.47) &99.79(0.35)\\
			14 &96.98(1.92) &98.31(0.71) &94.86(1.09) &99.81(0.23) &99.94(0.04) &98.86(0.46) &99.50(0.52)\\
			15 &94.27(2.92) &98.26(3.33) &98.96(0.36) &99.95(0.07) &99.68(0.15) &99.63(0.29) &100.00(0.01)\\
			16 &99.87(0.19) &99.99(0.04) &98.62(0.90) &100.00(0.00) &100.00(0.00) &100.00(0.00) &100.00(0.00)\\
			\midrule
			OA (\%) &98.69(0.50) &99.41(0.53) &99.08(0.11) &99.43(0.14) &\underline{99.65(0.06)} &99.25(0.29) &\textbf{99.93(0.03)}\\
			AA (\%) &98.89(0.25) &99.36(0.37) &98.85(0.05)  &\underline{99.76(0.05)} &99.42(0.13) &99.22(0.19) &\textbf{99.90(0.07)}\\
			$\kappa$ &98.54(0.55) &99.34(0.59) &98.97(0.12) &99.37(0.16) &\underline{99.61(0.07)} &99.17(0.33) &\textbf{99.93(0.03)}\\
			\bottomrule
	\end{tabular}}
\end{table*}

\begin{table*}[!t]
	\caption{Classification performance on the University of Pavia dataset. The highest value is bolded and the second-best is underlined.}
	\label{accuracy3}
	\vspace{1mm}
	\centering
	\resizebox{1.0\linewidth}{!}{ %
		\begin{tabular}{l|rrrrrr|r}
			\toprule        $\textmd{Class}$&$\textmd{IFRF\cite{IFRF17}}$&$\textmd{ARM\cite{ARM2017}}$&$\textmd{SGL\cite{SGL}}$&$\textmd{MSSGU\cite{MSSGU}}$&$\textmd{DMSGer\cite{DMSGer}}$&$\textmd{ConGCN\cite{ConGCN}}$&$\textmd{GWCL}$\\
			\midrule
			1 &85.15(7.06) &84.73(8.06) &88.26(3.06) &98.14(0.72) &93.37(2.24) &93.11(2.43) &98.74(1.45)\\
			2 &98.83(0.53) &97.33(1.05) &97.24(1.45) &82.90(2.30) &96.31(0.85) &96.55(1.87) &98.20(1.27)\\
			3 &83.46(4.72) &86.67(4.00) &94.19(2.64) &100.00(0.00) &99.91(0.11) &97.24(1.83) &95.63(2.34)\\
			4 &85.76(7.41) &89.59(5.46) &92.76(1.97) &98.44(0.45) &96.14(0.58) &93.91(1.83) &85.57(4.64)\\
			5 &99.50(0.74) &99.83(0.28) &99.30(0.82) &100.00(0.00) &99.47(0.15) &98.80(0.15) &99.61(0.12)\\
			6 &91.21(3.12) &93.40(4.38) &99.30(0.82) &100.00(0.00) &99.10(0.63) &100.00(0.00) &99.97(0.06)\\
			7 &84.24(6.12) &87.38(7.14) &99.31(0.23) &100.00(0.00) &99.77(0.33) &99.12(0.44) &99.29(1.16)\\
			8 &74.96(4.86) &78.25(6.29) &94.07(2.12) &99.48(0.19) &97.54(0.57) &94.76(1.93) &98.46(1.45)\\
			9 &66.16(12.01) &84.99(5.83) &99.67(0.03)  &99.86(0.11) &99.67(0.12) &82.81(3.20) &89.40(6.29)\\
			\midrule
			OA (\%) &90.10(1.57) &91.22(1.92) &95.58(0.60)  &92.06(1.01) &\underline{96.73(0.48)} &95.97(0.90)  &\textbf{97.38(0.60)}\\
			AA (\%) &85.48(1.86) &89.13(2.29) &96.07(0.47) &\underline{97.65(0.26)} &\textbf{97.92(0.31)} &95.14(0.57) &96.10(0.92)\\
			$\kappa$ &87.04(2.03) &88.45(2.46) &94.17(0.78) &89.81(1.26) &\underline{95.59(0.62)} &94.69(1.17) &\textbf{96.52(0.78)}\\
			\bottomrule
	\end{tabular} }
\end{table*}

(2) \textit{Results on Salinas}: For the Salinas dataset, as shown in Table~\ref{accuracy2}, all methods for comparison achieve over $98\%$ across the three metrics. However, GWCL further improves upon these results, attaining over $99.9\%$ accuracy with nearly the lowest standard deviations. In addition, we observed that GWCL achieved $100\%$ classification accuracy in nine classes, and even the worst-performing Class 14 achieved $99.50\%$ accuracy. Therefore, GWCL clearly outperforms all other methods, both overall and in individual class performance. The classification maps of different methods are shown in Fig.~\ref{samap} visually. It can be observed that GWCL is the closest to ground-truth and other methods have obvious misclassification areas. Especially for the classes represented by bluish gray, IFRF, ARM, and SGL all have large-area classification errors.

(3) \textit{Results on University of Pavia}: As shown in Table~\ref{accuracy3}, GWCL achieves the best OA and $\kappa$ and outperforms the second best OA by $0.65\%$, and the second best $\kappa$ by $1\%$ almost. In terms of AA, although GWCL does not outperform DMSGer and MSSGU due to its general performance in Class 4 and Class 9, it remains within an acceptable range. Out of the nine classes, GWCL achieved classification accuracy below $98\%$ in only three classes. In contrast, while MSSGU achieved over $98\%$ accuracy in eight classes, it performed extremely poorly in Class 2, which adversely affected its overall accuracy. Fig.~\ref{pumap} shows the classification map of various methods. Compared with the traditional methods IFRF and ARM, the misclassification areas of SGL, DMSGer and ConGCN that are based on superpixels are significantly smaller. Corresponding to Table~\ref{accuracy2}, MSSGU performs poorly in the class represented by light green. In contrast, GWCL performs significantly better in this large area land cover. This is due to the fact that GWCL makes full use of the spatial relationship between pixels.
\subsection{Parameter Analysis}
In this section, experiments are conducted to justify the selection of several key hyperparameters, including the dimension of pixel-spectral features, trade-off parameters in similarity calculation, and the number of neighbors in $K$-nearest neighbor graph.

\begin{figure}[!t]
	\centering
	\includegraphics[width=0.48\textwidth]{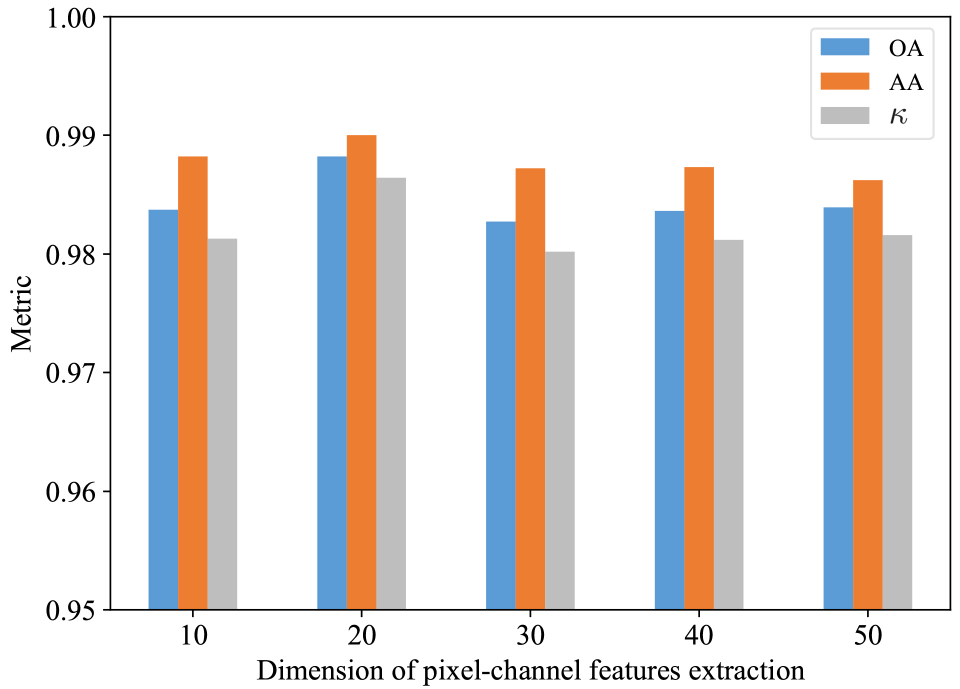}\\
	\caption{OA, AA, and $\kappa$ vary with $\beta$ on the Indian Pines dataset.}\label{dimension}
\end{figure}

Fig.~\ref{dimension} shows the OA, AA and $\kappa$ coefficient vary with dimensions of pixel-spectral features on the Indian Pines dataset. The raw HSI data is reduced to dimensions of 10, 20, 30, 40 and 50, respectively. It can be seen that when the dimension is 20, the OA, AA and $\kappa$ can achieve relatively high values. This is because dimensions lower than 20 may lead to information loss, while higher dimensions may introduce information redundancy. Surprisingly, the parameter had little effect on the classification accuracy. Within the selected numerical range, all three evaluation indices are almost above $98\%$.

\begin{figure}[!t]
	\centering	\includegraphics[width=0.48\textwidth]{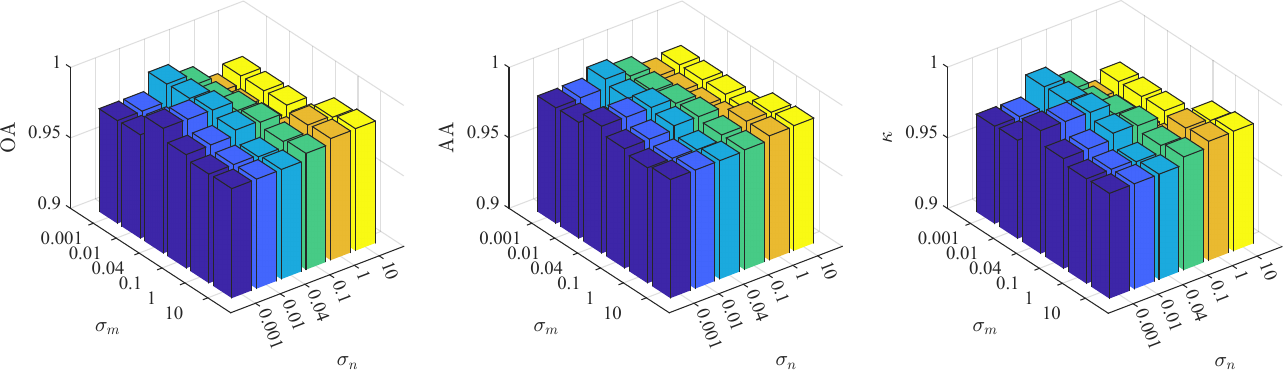}\\
	\caption{OA, AA, and $\kappa$ vary with $\sigma_m$ and $\sigma_n$ on the Indian Pines dataset.}
	\label{sigma}
\end{figure}

Bar charts in Fig.~\ref{sigma} shows the OA, AA, and $\kappa$ vary with trade-off parameters in similarity on the Indian Pines dataset. The $x$-axis and $y$-axis of bar charts represent $\sigma_m$ and $\sigma_n$ respectively. Specifically, for $\sigma_m$ and $\sigma_n$, we selected five values: 0.001, 0.01, 0.1, 1, and 10 to observe the changes in OA, AA, and $\kappa$, respectively. Moreover, experiments reveal that each index reach its maximum when $\sigma_m$ and $\sigma_n$ are set to 0.04 and 0.001, respectively. Therefore, Fig.~\ref{sigma} shows the variation of each metric for $\sigma_m$ and $\sigma_n$ across the aforementioned six values. It can be observed that the three indicators are not greatly affected by the two parameters, or even almost unaffected within a certain range. We have considered both spatial and spectral dimensions when constructing the graph, allowing the network to learn spatial relationships effectively. This is primarily reflected in the affinity graph, which captures spatial dependencies and helps improve classification performance.

\begin{figure}[H]
	\centering
	\includegraphics[width=0.42\textwidth]{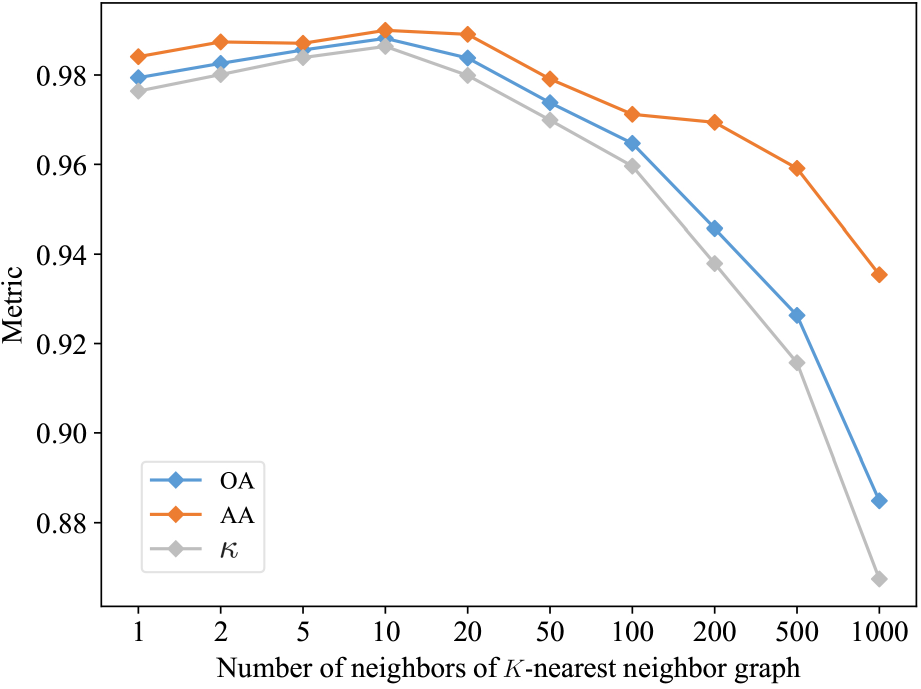}\\
	\caption{OA, AA, and $\kappa$ vary with $K$ on the Indian Pines dataset.}\label{k}
\end{figure}

Fig.~\ref{k} illustrates the impact of the number of neighbors on the Indian Pines dataset, where the number of neighbors is set to 1, 2, 5, 10, 20, 50, 100, 200, 500, and 1000. It demonstrates that 10 neighbors achieved the best OA, AA, and $\kappa$. In our weighted contrastive learning approach, the number of neighbors in the $K$-nearest neighbor graph determines the selection of negative sample pairs. Setting the number to 10 ensures that each sample has at least 10 similar samples to form positive pairs with it, while the remaining samples serve as negative examples. These 10 positive examples, weighted appropriately, have a more precise influence on the model parameters, leading to better feature learning. We conjecture that the best number of neighbors is influenced by the size of HSI and regional shape.

\subsection{Ablation Study}
Ablation experiments are conducted to assess the effectiveness of key components, including the proposed loss functions, training stages, and the pixel-spatial information embedded in the features. By systematically disabling components of the model, these experiments help determine their individual impact on overall performance.

Table~\ref{ablation} shows the results of ablation experiments on the Indian Pines dataset. It can be seen that OA, AA, and $\kappa$ decline when any one of the components is removed, especially the second training stage, semi-supervised weighted contrastive loss and pixel-spatial information embedded in features, which happen to be the core components where GWCL comes into play. Specifically, we find that after removing the semi-supervised weighted contrastive loss or the second training stage, the overall accuracy is reduced by more than $2\%$, indicating that GWCL played a leading role in improving the overall performance of the model.
\begin{table}[!t]
	\centering
	\caption{Ablation study results. The best results for OA, AA, and $\kappa$ are bolded.}\label{ablation}
	\resizebox{1.0\linewidth}{!}{
		\begin{tabular}{l| l|c c c}
			\toprule
			{Ablation} & {Settings} & {OA} & {AA} & $\kappa$ \\
			\midrule
			GWCL                  &no ablation   &\textbf{0.9882} &\textbf{0.9900} &\textbf{0.9864} \\
			\midrule
			Training stages  & w/o the first stage          &0.9819  &0.9877  &0.9793  \\
			&w/o the second stage         &0.9624  &0.9761  &0.9569  \\
			\midrule
			Losses           & w/o $\mathcal{L}_{\rm gwcl}$       &0.9673  &0.9793   &0.9626 \\
			&w/o $\mathcal{L}_{\rm ce}$       &0.9842  &0.9881    &0.9819\\
			\midrule
			Spatial information     & w/o spatial information      &0.9700 &0.9798  &0.9656 \\
			\midrule
	\end{tabular}}
\end{table}
\section{Conclusion}\label{sec:con}
This paper proposes GWCL for semi-supervised HSI classification. It constructs a pixel-level graph, avoids the superpixel segmentation step, and achieves mini-batch learning with reduced RAM requirements. The proposed semi-supervised weighted contrastive loss effectively extracts supervision information from a large number of unlabeled samples while fully utilizing labeled data. The superiority of the proposed method is demonstrated through comparison experiments on Indian Pines, Salinas, and University of Pavia datasets.  In the future, we plan to explore the use of graph neural networks for feature learning to enhance classification performance further. Additionally, we will investigate more contrastive learning theories to provide further guidance for improving classification results.
\section*{Disclosures}
The authors declare that there are no financial interests, commercial affiliations, or other potential conflicts of interest that could have influenced the objectivity of this research or the writing of this paper.
\section*{Code, Data, and Materials Availability}
The source code is available at \url{https://github.com/kunzhan/semiHSI}. The download links for the three datasets are provided in README.md of the GitHub repository.
\section*{Acknowledgments}
This work was supported by Key Laboratory of AI and Information Processing, Education Department of Guangxi Zhuang Autonomous Region (Hechi University) under No.~2024GXZDSY003.

{
\small
\bibliographystyle{ieeenat_fullname}
\bibliography{thisbib}
}



\end{document}


%% file: gwcl.bbl
\begin{thebibliography}{32}
\providecommand{\natexlab}[1]{#1}
\providecommand{\url}[1]{\texttt{#1}}
\expandafter\ifx\csname urlstyle\endcsname\relax
  \providecommand{\doi}[1]{doi: #1}\else
  \providecommand{\doi}{doi: \begingroup \urlstyle{rm}\Url}\fi

\bibitem[Achanta et~al.(2012)Achanta, Shaji, Smith, Lucchi, Fua, and
  Süsstrunk]{SLIC}
Radhakrishna Achanta, Appu Shaji, Kevin Smith, Aurelien Lucchi, Pascal Fua, and
  Sabine Süsstrunk.
\newblock Slic superpixels compared to state-of-the-art superpixel methods.
\newblock \emph{IEEE Transactions on Pattern Analysis and Machine
  Intelligence}, 34\penalty0 (11):\penalty0 2274--2282, 2012.

\bibitem[Barber and Agakov(2003)]{IBA}
David Barber and Felix~V. Agakov.
\newblock The {IM} algorithm: A variational approach to information
  maximization.
\newblock In \emph{NeurIPS}, 2003.

\bibitem[Belghazi et~al.(2018)Belghazi, Baratin, Rajeshwar, Ozair, Bengio,
  Courville, and Hjelm]{MINE}
Mohamed~Ishmael Belghazi, Aristide Baratin, Sai Rajeshwar, Sherjil Ozair,
  Yoshua Bengio, Aaron Courville, and Devon Hjelm.
\newblock Mutual information neural estimation.
\newblock In \emph{ICML}, pages 531--540, 2018.

\bibitem[Camps-Valls et~al.(2007)Camps-Valls, Bandos~Marsheva, and
  Zhou]{4305352}
Gustavo Camps-Valls, Tatyana~V. Bandos~Marsheva, and Dengyong Zhou.
\newblock Semi-supervised graph-based hyperspectral image classification.
\newblock \emph{IEEE Transactions on Geoscience and Remote Sensing},
  45\penalty0 (10):\penalty0 3044--3054, 2007.

\bibitem[Chung(1997)]{chung1997spectral}
Fan~RK Chung.
\newblock \emph{Spectral Graph Theory}.
\newblock American Mathematical Soc., 1997.

\bibitem[Cohen(1960)]{cohen1960coefficient}
J. Cohen.
\newblock A coefficient of agreement for nominal scales.
\newblock \emph{Educational and Psychological Measurement}, 20\penalty0
  (1):\penalty0 37--46, 1960.

\bibitem[Foody(2002)]{foody2002status}
G.~M. Foody.
\newblock Status of land cover classification accuracy assessment.
\newblock \emph{Remote Sensing of Environment}, 80\penalty0 (1):\penalty0
  185--201, 2002.

\bibitem[Grill et~al.(2020)Grill, Strub, Altch{\'e}, Tallec, Richemond,
  Buchatskaya, Doersch, Avila~Pires, Guo, Gheshlaghi~Azar, et~al.]{BYOL}
Jean-Bastien Grill, Florian Strub, Florent Altch{\'e}, Corentin Tallec, Pierre
  Richemond, Elena Buchatskaya, Carl Doersch, Bernardo Avila~Pires, Zhaohan
  Guo, Mohammad Gheshlaghi~Azar, et~al.
\newblock Bootstrap your own latent: A new approach to self-supervised
  learning.
\newblock In \emph{NeurIPS}, pages 21271--21284, 2020.

\bibitem[Han et~al.(2024)Han, Tian, Xia, and Zhan]{infomatch}
Qi Han, Zhibo Tian, Chengwei Xia, and Kun Zhan.
\newblock {InfoMatch}: Entropy neural estimation for semi-supervised image
  classification.
\newblock In \emph{IJCAI}, pages 4089--4097, 2024.

\bibitem[Hong et~al.(2021)Hong, Gao, Yao, Zhang, Plaza, and Chanussot]{miniGCN}
Danfeng Hong, Lianru Gao, Jing Yao, Bing Zhang, Antonio Plaza, and Jocelyn
  Chanussot.
\newblock Graph convolutional networks for hyperspectral image classification.
\newblock \emph{IEEE Transactions on Geoscience and Remote Sensing},
  59\penalty0 (7):\penalty0 5966--5978, 2021.

\bibitem[Jia et~al.(2024)Jia, Jiang, Zhang, Xu, and Jia]{GIG}
Sen Jia, Shuguo Jiang, Shuyu Zhang, Meng Xu, and Xiuping Jia.
\newblock Graph-in-graph convolutional network for hyperspectral image
  classification.
\newblock \emph{IEEE Transactions on Neural Networks and Learning Systems},
  35\penalty0 (1):\penalty0 1157--1171, 2024.

\bibitem[Jiang et~al.(2019)Jiang, Ma, Wang, Chen, and Liu]{RLPA}
Junjun Jiang, Jiayi Ma, Zheng Wang, Chen Chen, and Xianming Liu.
\newblock Hyperspectral image classification in the presence of noisy labels.
\newblock \emph{IEEE Transactions on Geoscience and Remote Sensing},
  57\penalty0 (2):\penalty0 851--865, 2019.

\bibitem[Jiang et~al.(2022)Jiang, Ma, and Liu]{MSSG}
Junjun Jiang, Jiayi Ma, and Xianming Liu.
\newblock Multilayer spectral–spatial graphs for label noisy robust
  hyperspectral image classification.
\newblock \emph{IEEE Transactions on Neural Networks and Learning Systems},
  33\penalty0 (2):\penalty0 839--852, 2022.

\bibitem[Kang et~al.(2014)Kang, Li, and Benediktsson]{IFRF17}
Xudong Kang, Shutao Li, and Jón~Atli Benediktsson.
\newblock Feature extraction of hyperspectral images with image fusion and
  recursive filtering.
\newblock \emph{IEEE Transactions on Geoscience and Remote Sensing},
  52\penalty0 (6):\penalty0 3742--3752, 2014.

\bibitem[Liu et~al.(2011)Liu, Tuzel, Ramalingam, and Chellappa]{ERS}
Ming-Yu Liu, Oncel Tuzel, Srikumar Ramalingam, and Rama Chellappa.
\newblock Entropy rate superpixel segmentation.
\newblock In \emph{CVPR 2011}, pages 2097--2104, 2011.

\bibitem[Liu et~al.(2022)Liu, Xiao, Yang, and Wei]{MSSGU}
Qichao Liu, Liang Xiao, Jingxiang Yang, and Zhihui Wei.
\newblock Multilevel superpixel structured graph u-nets for hyperspectral image
  classification.
\newblock \emph{IEEE Transactions on Geoscience and Remote Sensing},
  60:\penalty0 1--15, 2022.

\bibitem[Ma et~al.(2023)Ma, Zhang, Zhang, and Zhan]{MILBO}
Yixuan Ma, Xiaolin Zhang, Peng Zhang, and Kun Zhan.
\newblock Entropy neural estimation for graph contrastive learning.
\newblock In \emph{ACM Multimedia}, pages 435--443, 2023.

\bibitem[Mou et~al.(2020)Mou, Lu, Li, and Zhu]{Nonlocal}
Lichao Mou, Xiaoqiang Lu, Xuelong Li, and Xiao~Xiang Zhu.
\newblock Nonlocal graph convolutional networks for hyperspectral image
  classification.
\newblock \emph{IEEE Transactions on Geoscience and Remote Sensing},
  58\penalty0 (12):\penalty0 8246--8257, 2020.

\bibitem[Nguyen et~al.(2010)Nguyen, Wainwright, and Jordan]{Nguyen}
XuanLong Nguyen, Martin~J. Wainwright, and Michael~I. Jordan.
\newblock Estimating divergence functionals and the likelihood ratio by convex
  risk minimization.
\newblock \emph{IEEE Transactions on Information Theory}, 56\penalty0
  (11):\penalty0 5847--5861, 2010.

\bibitem[Nowozin et~al.(2016)Nowozin, Cseke, and Tomioka]{fgan}
Sebastian Nowozin, Botond Cseke, and Ryota Tomioka.
\newblock $f$-{GAN}: Training generative neural samplers using variational
  divergence minimization.
\newblock In \emph{NeurIPS}, 2016.

\bibitem[Oord et~al.(2018)Oord, Li, and Vinyals]{CPC}
Aaron van~den Oord, Yazhe Li, and Oriol Vinyals.
\newblock Representation learning with contrastive predictive coding.
\newblock \emph{arXiv preprint arXiv:1807.03748}, 2018.

\bibitem[Sellars et~al.(2020)Sellars, Aviles-Rivero, and Schonlieb]{SGL}
Philip Sellars, Angelica~I. Aviles-Rivero, and Carola-Bibiane Schonlieb.
\newblock Superpixel contracted graph-based learning for hyperspectral image
  classification.
\newblock \emph{IEEE Transactions on Geoscience and Remote Sensing},
  58\penalty0 (6):\penalty0 4180--4193, 2020.

\bibitem[Su et~al.(2022)Su, Jiang, Gao, You, Sun, and Li]{graphcut}
Yuanchao Su, Mengying Jiang, Lianru Gao, Xueer You, Xu Sun, and Pengfei Li.
\newblock Graph-cut-based node embedding for dimensionality reduction and
  classification of hyperspectral remote sensing images.
\newblock In \emph{IGARSS}, pages 1720--1723, 2022.

\bibitem[Tian et~al.(2023)Tian, Zhang, Zhang, and Zhan]{DSSN2023}
Zhibo Tian, Xiaolin Zhang, Peng Zhang, and Kun Zhan.
\newblock Improving semi-supervised semantic segmentation with dual-level
  {S}iamese structure network.
\newblock In \emph{ACM Multimedia}, pages 4200--4208, 2023.

\bibitem[Wan et~al.(2021)Wan, Gong, Zhong, Pan, Li, and Yang]{CAD_GCN}
Sheng Wan, Chen Gong, Ping Zhong, Shirui Pan, Guangyu Li, and Jian Yang.
\newblock Hyperspectral image classification with context-aware dynamic graph
  convolutional network.
\newblock \emph{IEEE Transactions on Geoscience and Remote Sensing},
  59\penalty0 (1):\penalty0 597--612, 2021.

\bibitem[Wu et~al.(2024)Wu, Al-qaness, Al-Alimi, Dahou, Abd~Elaziz, and
  Ewees]{wu2024hyperspectral}
Guoyong Wu, Mohammed~AA Al-qaness, Dalal Al-Alimi, Abdelghani Dahou, Mohamed
  Abd~Elaziz, and Ahmed~A Ewees.
\newblock Hyperspectral image classification using graph convolutional network:
  A comprehensive review.
\newblock \emph{Expert Systems with Applications}, 257:\penalty0 125106, 2024.

\bibitem[Yang et~al.(2024)Yang, Tang, Zhang, Ma, Liu, Jia, and Jiao]{DMSGer}
Yuqun Yang, Xu Tang, Xiangrong Zhang, Jingjing Ma, Fang Liu, Xiuping Jia, and
  Licheng Jiao.
\newblock Semi-supervised multiscale dynamic graph convolution network for
  hyperspectral image classification.
\newblock \emph{IEEE Transactions on Neural Networks and Learning Systems},
  35\penalty0 (5):\penalty0 6806--6820, 2024.

\bibitem[Yu et~al.(2023)Yu, Wan, Li, Yang, and Gong]{ConGCN}
Wentao Yu, Sheng Wan, Guangyu Li, Jian Yang, and Chen Gong.
\newblock Hyperspectral image classification with contrastive graph
  convolutional network.
\newblock \emph{IEEE Transactions on Geoscience and Remote Sensing},
  61:\penalty0 1--15, 2023.

\bibitem[Yu and Zhang(2023)]{MView}
Xiao Yu and Qiang Zhang.
\newblock Graph-based multi-view learning for hyperspectral remote sensing
  image classification.
\newblock In \emph{IGARSS}, pages 7222--7225, 2023.

\bibitem[Zhan et~al.(2017)Zhan, Wang, Xie, Zhang, and Min]{ARM2017}
Kun Zhan, Haibo Wang, Yuange Xie, Chutong Zhang, and Yufang Min.
\newblock Albedo recovery for hyperspectral image classification.
\newblock \emph{Journal of Electronic Imaging}, 26\penalty0 (4):\penalty0
  043010, 2017.

\bibitem[Zhou et~al.(2003)Zhou, Bousquet, Lal, Weston, and
  Sch\"{o}lkopf]{NIPS2003_87682805}
Dengyong Zhou, Olivier Bousquet, Thomas Lal, Jason Weston, and Bernhard
  Sch\"{o}lkopf.
\newblock Learning with local and global consistency.
\newblock In \emph{NeurIPS}, 2003.

\bibitem[Zhu et~al.(2022)Zhu, Zhao, Qin, and Feng]{SLGConv}
Wenxiang Zhu, Chunhui Zhao, Boao Qin, and Shou Feng.
\newblock Short and long range graph convolution network for hyperspectral
  image classification.
\newblock In \emph{IGARSS}, pages 3564--3567, 2022.

\end{thebibliography}
